\documentclass[conference]{IEEEtran}
\usepackage{cite}
\usepackage{amsmath,amssymb,amsfonts}
\usepackage{graphicx}
\usepackage{textcomp}
\usepackage{xcolor}
\usepackage{caption}
\usepackage{subcaption}
\usepackage{multirow}
\usepackage{amssymb}
\usepackage{amsmath}
\usepackage{url}
\usepackage{soul}
\usepackage{color}
\usepackage{booktabs}
\usepackage{enumitem}
\usepackage{algorithm}
\usepackage[noend]{algpseudocode}

\def\BibTeX{{\rm B\kern-.05em{\sc i\kern-.025em b}\kern-.08em
    T\kern-.1667em\lower.7ex\hbox{E}\kern-.125emX}}

\begin{document}

\title{Learning~Private~Neural~Language~Modeling\\~with~Attentive~Aggregation}

\author{\IEEEauthorblockN{Shaoxiong Ji\IEEEauthorrefmark{1}, Shirui Pan\IEEEauthorrefmark{2}, Guodong Long\IEEEauthorrefmark{3}, Xue Li\IEEEauthorrefmark{1}, Jing Jiang\IEEEauthorrefmark{3}, Zi Huang\IEEEauthorrefmark{1}}
\IEEEauthorblockA{\IEEEauthorrefmark{1} School of ITEE, Faculty of EAIT, The University of Queensland, Australia \\ \IEEEauthorrefmark{2} Faculty of Information Technology, Monash University, Australia \\
\IEEEauthorrefmark{3} Centre for Artificial Intelligence, Faculty of Engineering and IT, University of Technology Sydney, Australia\\
Email: shaoxiong.ji@uq.edu.au, shirui.pan@monash.edu, guodong.long@uts.edu.au, \\xueli@itee.uq.edu.au, jing.jiang@uts.edu.au, huang@itee.uq.edu.au}
}

\maketitle

\begin{abstract}
Mobile keyboard suggestion is typically regarded as a word-level language modeling problem. Centralized machine learning techniques require the collection of massive user data for training purposes, which may raise privacy concerns in relation to users' sensitive data. 
Federated learning (FL) provides a promising approach to learning private language modeling for intelligent personalized keyboard suggestions by training models on distributed clients rather than training them on a central server. To obtain a global model for prediction, existing FL algorithms simply average the client models and ignore the importance of each client during model aggregation. Furthermore, there is no optimization for learning a well-generalized global model on the central server. 
To solve these problems, we propose a novel model aggregation with an attention mechanism considering the contribution of client models to the global model, together with an optimization technique during server aggregation. 
Our proposed attentive aggregation method minimizes the weighted distance between the server model and client models by iteratively updating parameters while attending to the distance between the server model and client models.
Experiments on two popular language modeling datasets and a social media dataset show that our proposed method outperforms its counterparts in terms of perplexity and communication cost in most settings of comparison.

\end{abstract}


\section{Introduction}
\label{sec::intro}
With the advances in mobile technology, smart phones and wearable devices like smart watches are becoming increasingly popular in modern life. A study conducted in 2017 shows that the majority of participants spent five hours or more every day on their smartphones\footnote{Reported by The Statistics Portal available at \url{https://www.statista.com/statistics/781692/worldwide-daily-time-spent-on-smartphone/}, retrieved in Dec, 2018}. These mobile devices generate a massive amount of distributed data such as text messages, travel trajectories and health status. 

In the traditional machine learning approaches, cloud-based servers send a data collection request to the clients, collect data from clients, train a centralized model, and make predictions for clients. However, this centralized way of learning relies heavily on the central server. Moreover, there are privacy issues, especially in relation to sensitive data, if the central server is hacked into or misused by other third parties. 

Recently, a distributed learning technique called federated learning has attracted great interest from the research community \cite{mcmahan2017communication, geyer2017differentially, chen2018federated} under the umbrella of distributed machine learning. It protects the privacy of data by learning a shared model by distributed training on local client devices without collecting the data on a central server. Distributed intelligent agents take a shared global model from the central server's parameters as initialization to train their own private models using personal data, and make predictions on their own physical devices. There are many applications of federated learning in the real world, for example, predicting the most likely photos a mobile user would like to share on the social websites \cite{kim2016predicting}, predicting the next word for mobile keyboards \cite{arnold2016suggesting}, retrieving the most important notifications, and detecting the spam messages \cite{he2017new}.

Of these applications, mobile keyboard suggestion as a language modeling problem is one of the most common tasks because it involves with user interaction which can give instant labeled data for supervised learning. In practice, the mobile keyboard applications predict the next word with several options when a user is typing a sentence. 
Generally speaking, an individual's language usage expresses that person's language preference and patterns. With the recent advances in deep neural networks, a language model combined with neural networks, called neural language modeling, has been developed. Of these neural network models, recurrent neural networks (RNNs) which capture the temporal relations in sentences has significantly improved the field of language modeling. Specific RNNs include long short-term memory (LSTM) \cite{hochreiter1997long} and its variants such as gated recurrent unit (GRU) \cite{cho2014learning}.

In the real-world scenario, users' language input and preferences are sensitive and may contain some private content including private personal profiles, financial records, passwords, and social relations. Thus, to protect the user's privacy, a federated learning technique with data protection is a promising solution. In this paper, we take this application as learning word-level private neural language modeling for each user. 

Federated learning learns a shared global model by the aggregation of local models on client devices. But the original paper on federated learning \cite{mcmahan2017communication} only uses a simple average on client models, taking the number of samples in each client device as the weight of the average. 
In the mobile keyboard applications, language preferences may vary from individual to individual. The contributions of client language models to the central server are quite different.
To learn a generalized private language model that can be quickly adapted to different people's language preferences, knowledge transferring between server and client, especially the well-trained clients models, should be considered. 

In this paper, we introduce an attention mechanism for model aggregation. It is proposed to automatically attend to the weights of the relation between the server model and different client models. The attentive weights are then taken to minimize the expected distance between the server model and client models. The advantages of our proposed method are: 1) it considers the relation between the server model and client models and their weights, and 2) it optimizes the distance between the server model and client models in parameter space to learn a well-generalized server model.

Our contributions in this paper are as follows:
\begin{itemize}
\item Our work first introduces the attention mechanism to aggregate multiple distributed models. The proposed attentive aggregation can be further applied to improve broad methods and applications using distributed machine learning.
\item In the server optimization, we propose a novel layer-wise soft attention to capturing the ``attention'' among many local models' parameters.
\item As demonstrated by the experiments on private neural language modeling task for mobile keyboard suggestions, the proposed method achieves a comparable performance in terms of perplexity and communication rounds.
\end{itemize}

The structure of this paper is organized as follows. In Section \ref{sec:related}, related works including federated learning, attention mechanism and neural language modeling are reviewed. Our proposed attentive federated aggregation is introduced in Section \ref{sec:method}. Experimental settings and results are given in Section \ref{sec:exp} together with the comparison and analysis. In Section \ref{sec:conclusion}, the conclusion is drawn. 

\section{Related~Work}
\label{sec:related}
This paper relates to federated learning, language modeling, and attention mechanism.

\subsection{Federated~Learning~}
Federated learning is proposed by McMahan et al. \cite{mcmahan2017communication} to decouple training procedures from data collection by an iterative averaging model. It can perform distributed training and communication-efficient learning from decentralized data to achieve the goal of privacy preservation. Geyer et al. proposed a differential privacy-preserving technique on the client side to balance the performance and privacy \cite{geyer2017differentially}. Popov et al. proposed the fine-tuning of distributed private data to learn language models \cite{popov2018distributed}. 
The federated learning technique is useful in many fields. Chen et al. combined federated learning with meta learning for recommendation \cite{chen2018federated}. Kim et al. proposed federated tensor factorization to discover clinical concepts from electronic health records \cite{kim2017federated}. The federated setting can also be integrated into other machine learning settings. 
Smith et al. \cite{smith2017federated} proposed a framework that fits well with multi-task learning and federated setting to tackle the statistical challenge of distributed machine learning.

Communication efficiency is one of the performance metrics for federated learning techniques. To improve communication efficiency, Kone\v{c}n\`{y} et al. proposed structured updates and sketched updates to reduce the uplink communication costs \cite{konevcny2016federated}. 
It is also studied under the umbrella of distributed machine learning. Alistarch et al. proposed quantized compression and the encoding of stochastic gradient descent \cite{alistarh2017qsgd} to achieve efficient communication.
Wen et al. used ternary gradients to reduce the communication cost \cite{wen2017terngrad}.

\subsection{Neural~Language~Modeling~} Language modeling, as one of the most crucial natural language processing tasks, has achieved better performance than classical methods using popular neural networks. Mikolov et al. used a simple recurrent neural network-based language model for speech recognition \cite{mikolov2010recurrent}. Recently, LSTM-based neural language models were developed to learn context over long sequences in large datasets \cite{jozefowicz2016exploring}. To facilitate learning in a language model and reduce the number of trainable parameters, Inan et al. proposed tying word vectors and word classifiers \cite{inan2016tying}. Press and Wolf evaluated the effect of weight tying and introduced a new regularization on the output embedding \cite{press2017using}.

\subsection{Attention~Mechanism~}
The attention mechanism is simply a vector serving to orient perception. It first became popular in the field of computer vision. Mnih et al. used it in recurrent neural network models for image classification \cite{mnih2014recurrent}. It was then widely applied in sequence-to-sequence natural language processing tasks like neural machine translation \cite{bahdanau2014neural}. Luong et al. extended attention-based RNNs and proposed two new mechanisms, i.e., the global attention and local attention \cite{luong2015effective}. Also, the attention mechanism can be used in convolutional models for sentence encoding like ABCNN for modeling sentence pairs \cite{yin2016abcnn}. Yang et al. proposed hierarchical attention networks for document classification \cite{yang2016hierarchical}. Shen et al. proposed directional self-attention for language understanding \cite{shen2017disan}.

\section{Proposed~Method}
\label{sec:method}
In this section, we firstly introduce the preliminaries of the federated learning framework, and then propose our attentive federated optimization algorithm to improve the generalizability for distributed clients by learning the attentive weights of selected clients during model aggregation. As for the client learner, we apply the gated recurrent unit (GRU) \cite{cho2014learning} as the client model for language modeling. Furthermore, we add a randomized mechanism \cite{geyer2017differentially} for learning differentially private client model.

\subsection{Preliminaries~of~Federated~Learning}
Federated learning decouples the model training and data collection \cite{mcmahan2017communication}. To learn a well generalized model, it uses model aggregation on the server side, which is similar to the works on meta-learning by learning a good initialization for quick adaptation \cite{finn2017model, nichol2018firstorder} and transfer learning by transferring knowledge between domains \cite{pan2010survey}. The basic federated learning framework comprises two main parts, i.e., server optimization in Algorithm \ref{alg:server} and local training in Algorithm \ref{alg:local} \cite{mcmahan2017communication}.

\vspace{0.1cm}
\noindent \textbf{Central~Model~Update.}
The server firstly chooses a client learning model and initializes the parameters of the client learner. It sets the fraction of the clients. Then, it waits for online clients for local model training. Once the selected number of clients finishes the model update, it receives the updated parameters and performs the server optimization. The parameter sending and receiving consists of one round of communication. Our proposed optimization is conducted in Line 9 of Algorithm \ref{alg:server}.

\begin{algorithm}
\caption{Optimization~for~Federated~Learning~on~Central~Server}
\label{alg:server}
\begin{algorithmic}[1]
\State K is the total number of clients; C is the client fraction; U is a set of all clients.
\State \textbf{Input}: server parameters $\theta_{t}$ at~$t$, client parameters $\theta^1_{t+1}, ..., \theta^m_{t+1}$ at $t+1$.
\State \textbf{Output}: aggregated server parameters $\theta_{t+1}$.
\Procedure{Server Execution}{} \Comment{Run on the server}
\State initialize~$\theta_0$

\For{each~round~$t$=1,~2,~...~}
\State $m\gets max(C \cdot K, 1)$
\State $S_t \gets \{u_i~|~u_i \in U\}_1^m $\Comment{Random~set~of~clients}
	\For{each~client~k~$\in~S_t$~on~local~device}
		\State $\theta_{t+1}^k \gets \text{ClientUpdate}(k,\theta_t)$
	\EndFor
	\State $\theta_{t+1} \gets \text{ServerOptimization}(\theta_t, \theta^k_{t+1})$
\EndFor
\EndProcedure
\end{algorithmic}
\end{algorithm}

\vspace{0.1cm}
\noindent \textbf{Private~Model~Update.}
Each online selected client receives the server model and performs secure local training on their own devices. For the neural language modeling, stochastic gradient descent is performed to update their GRU-based client models which is introduced in Section \ref{subsec:gru}. After several epochs of training, the clients send the parameters of their models to the central server over a secure connection. During this local training, user data can be stored on their own devices. 

\begin{algorithm}
\caption{Secure Local Training on Client}
\label{alg:local}
\begin{algorithmic}[1]
\State B is the local mini-batch size; E is the number of local epochs; $\beta$ is the momentum term; $\eta$ is the learning rate.
\State \textbf{Input}: ordinal of user $k$, user data $X$.
\State \textbf{Output}: updated user parameters $\theta_{t+1}$ at $t+1$.
\Procedure{Client Update}{$k$,~$\theta$} \Comment{Run on the $k$-th client}

\State $\textrm{B} \gets (split~user~data~X~into~batches)$
\For{each~local~epoch~i~from~1~to~E}
	\For{batch~b~$\in~\textrm{B}$}
		\State $ z_{t+1} \gets \beta z_t + \nabla\textit{L}(\theta_t)$	
		\State $ \theta_{t+1} \gets \theta_t - \eta z_{t+1} $
	\EndFor
\EndFor
\State send~$\theta_{t+1}$~to~server
\EndProcedure
\end{algorithmic}
\end{algorithm}

\subsection{Attentive~Federated~Aggregation}
\label{sub::attention}

\begin{figure}[htbp]
\begin{center}
\includegraphics[width=0.35\textwidth]{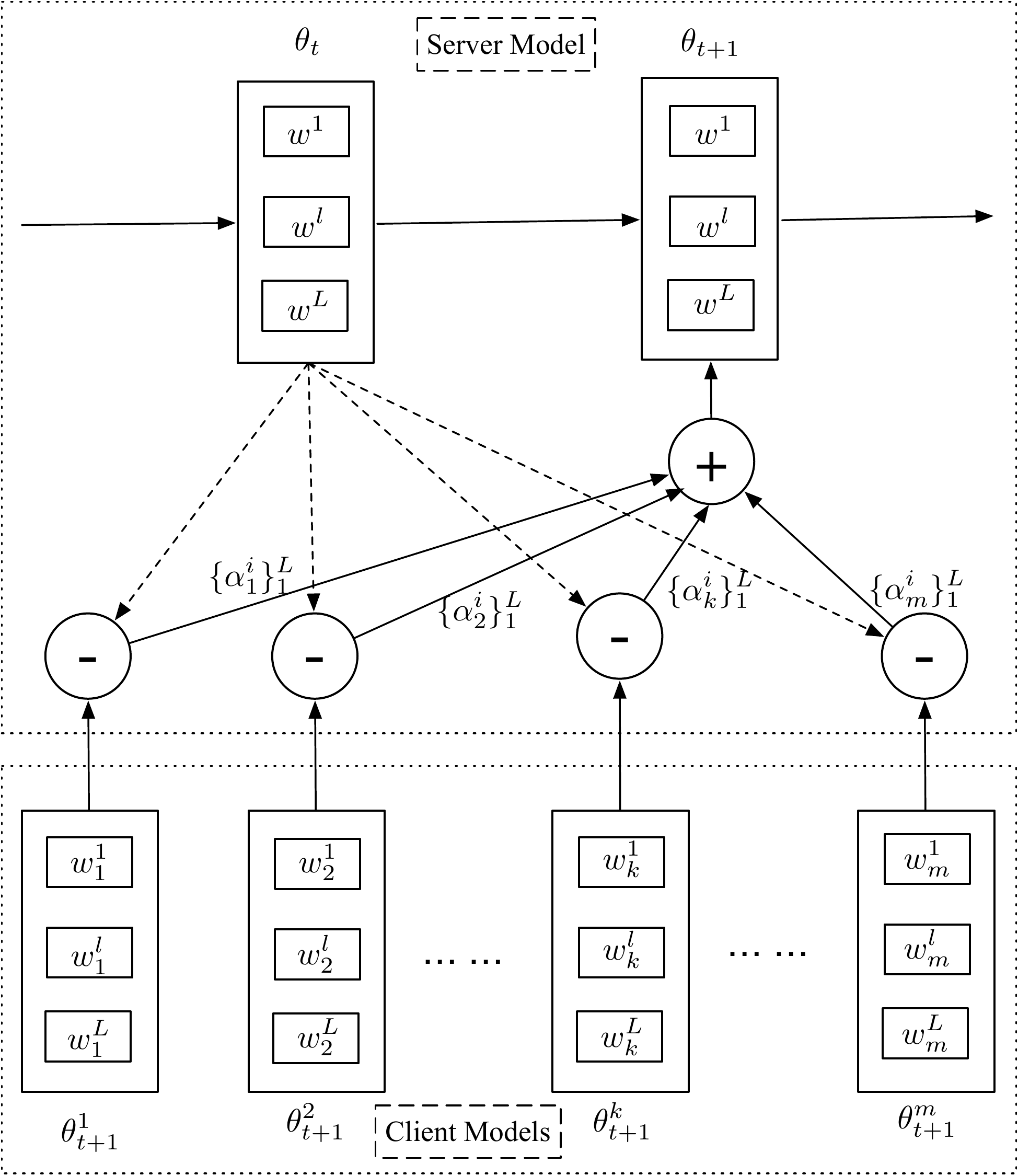}
\caption{The illustration of our proposed layer-wise attentive federated aggregation}
\label{fig:method}
\end{center}
\end{figure}

The most important part of federated learning is the federated optimization on the server side which aggregates the client models. In this paper, a novel federated optimization strategy is proposed to learn federated learning from decentralized client models. We call this Attentive Federated Aggregation, or \texttt{FedAtt} for short. It firstly introduces the attention mechanism for federated aggregation by aggregating the layer-wise contribution of neural language models of selected clients to the global model in the central server. An illustration of our proposed layer-wise attentive federated aggregation is shown in Figure \ref{fig:method} where the lower box represents the distributed client models and the upper box represents the attentive aggregation in the central server. The distributed client models in the lower box contain several neural layers. The notations of ``$\oplus$'' and ``$\ominus$'' stand for the layer-wise operation on the parameters of neural models. This illustration shows only a single time step. The federated updating uses our proposed attentive aggregation block to update the global model by iteration.

The intuition behind the federated optimization is to find an optimal global model that can generalize the client models well. In our proposed optimization algorithm, we take it as finding an optimal global model that is close to the client models in parameter space while considering the importance of selected client models during aggregation. The optimization objective is defined as 

\begin{equation}
\label{eq:obj}
\mathop{\arg \min}_{\theta_{t+1}}\sum_{k=1}^m[\frac{1}{2} \alpha_k L(\theta_t, \theta^k_{t+1})^2],
\end{equation}
where $\theta_t$ is the parameters of the global server model at time stamp $t$, $\theta^k_{t+1}$ is the parameters of the $k$-th client model at time stamp $t+1$, $L(\cdot,\cdot)$ is defined as the distance between two sets of neural parameters, and $\alpha_k$ is the attentive weight to measure the importance of weights for the client models. 
The objective is to minimize the weighted distance between server model and client models by taking a set of self-adaptive scores as the weights.

To attend the importance of client models, we propose a novel layer-wise attention mechanism on the parameters of the neural network models. The attention mechanism in this paper is quite similar to the soft attention mechanism. Unlike the popular attention mechanism applied to the data flow, our proposed attentive aggregation is applied on the learned parameters of each layer of the neural language models. We take the server parameters as a query and the client parameters as keys, and calculate the attention score in each layer of the neural networks. 

Given the parameters in the $l$-th layer of the server model denoted as $w^l$ and parameters in the $l$-th layer of the $k$-th client model denoted as $w^l_k$, the similarity between the query and the key in the $l$-th layer is calculated as the norm of the difference between two matrices, which is denoted as:
$$ s^l_k = \| w^l-w^l_k\|_p. $$ 
Then, we apply softmax on the similarity to calculate the layer-wise attention score for the $k$-th client in Equation \ref{eq:attention}. 
\begin{equation}
\label{eq:attention}
\alpha^l_k = softmax(s^l_k) = \frac{e^{s^l_k}}{\sum_{k=1}^m e^{s^l_k}}
\end{equation}
Our proposed attention mechanism on the parameters is layer-wise. There are attention scores for each layer in the neural networks. For each model, the attention score is $\alpha_k = \{\alpha^0_k, \alpha^1_k, \dots, \alpha^l_k, \dots \}$ in a non-parameter way.

Using the Euclidean distance for $L(\cdot,\cdot)$ and taking the derivative of the objective function in Equation \ref{eq:obj}, we get the gradient in the form of Equation \ref{eq:gradient}.

\begin{equation}
    \label{eq:gradient}
    \nabla = \sum_{k=1}^m\alpha_k(\theta_t-\theta_{t+1}^k)
\end{equation}
For the selected group of $m$ clients, we perform gradient descent to update the parameters of the global model in Equation \ref{eq:gd} as
\begin{equation}
\label{eq:gd}
\theta_{t+1} \gets \theta_t - \epsilon \sum_{k=1}^m\alpha_k(\theta_t-\theta_{t+1}^k),
\end{equation}
where $\epsilon$ is the step size. The full procedure of our proposed optimization algorithm is described in Algorithm \ref{alg:attentive}. It takes the server parameters $\theta_t$ at time stamp $t$ and client parameters $\theta^1_{t+1}, ..., \theta^m_{t+1}$ at time stamp $t+1$, and returns the updated parameters of the global server.

\begin{algorithm}
\caption{Attentive~Federated~Optimization}
\label{alg:attentive}
\begin{algorithmic}[1]
\State $k$ is the ordinal of clients; $l$ is the ordinal of neural layers; $\epsilon$ is the stepsize of server optimization
\State \textbf{Input}: server parameters $\theta_{t}$ at~$t$, client parameters $\theta^1_{t+1}, ..., \theta^m_{t+1}$ at $t+1$.
\State \textbf{Output}: aggregated server parameters $\theta_{t+1}$.
\Procedure{Attentive~Optimization}{$\theta_t$, $\theta^k_{t+1}$}
\State Initialize $\alpha = \{\alpha_1, \dots, \alpha_k, \dots, \alpha_m\}$ \Comment{attention for each clients}

\For{each~layer~$l=1,~2, ...~$}
	\For{each~user~k~}
		\State $ s^l_k = \| w^l-w^l_k\|_p $
	\EndFor
	\State $\alpha^l_k = softmax(s^l_k) = \frac{e^{s^l_k}}{\sum_{k=1}^m e^{s^l_k}}$
\EndFor
\State $\alpha_k = \{\alpha^0_k, \alpha^1_k, \dots, \alpha^l_k, \dots\}$
\State $\theta_{t+1} \gets \theta_t - \epsilon \sum_{k=1}^m\alpha_k(\theta_t-\theta_{t+1}^k)$
\State return $\theta_{t+1}$
\EndProcedure
\end{algorithmic}
\end{algorithm}

The advantage of our proposed layer-wise attentive federated aggregation and its optimization is as follows: 1) The aggregation of client models is fine-grained on each layer of the neural models considering the similarity between the client model and the server model in the parameter space. The learned features of each client model can be effectively selected to to produce a fine-tuned global server model. 2) By minimizing the expected distance between the client model and the server model, the learned server model is close to the client models in the parameter space and can well represent the federated clients.

\subsection{GRU-based~Client~Model}
\label{subsec:gru}
The learning process on the client side is model-agnostic. For different tasks, we can choose appropriate models in specific situations. In this paper, we use the gated recurrent unit (GRU) \cite{cho2014learning} for learning the language modeling on the client side. The GRU is a well-known and simpler variant of the Long Short-Term Memory (LSTM) \cite{hochreiter1997long}, by merging the forget gate and the input gate into a single gate as well as the cell state and the hidden state. In the GRU-based neural language model, words or tokens are firstly embedded into word vectors denoted as $X=\{x_0, x_1, \dots, x_t, \dots \}$ and then put into the recurrent loops. The calculation inside the recurrent module is as follows:
\begin{eqnarray*}
z_t &=& \sigma (w_z\cdot [h_{t-1},~x_t]),\\
r_t &=& \sigma (w_r \cdot [h_{t-1},~x_t]),\\
\tilde{h_t} &=& tanh (w\cdot [r_t*h_{t-1},~x_t]),\\
h_t &=& (1-z_t)*h_{t-1} + z_t* \tilde{h_t},
\end{eqnarray*}
where $z_t$ is the update gate, $r_t$ is the reset gate, $h_t$ is the hidden state, and $\tilde{h_t}$ is a new hidden state.

\subsection{Differential~Privacy}
To protect the client's data from an inverse engineering attack, we apply the randomized mechanism into federated learning \cite{geyer2017differentially}. This ensures differential privacy on the client side without revealing the client's data \cite{geyer2017differentially}.  This differentially private randomization was firstly proposed to apply on the federated averaging, where a white noise with the mean of $0$ and the standard deviation of $\sigma$ is added to the client parameters in Equation \ref{eq:dp-avg}. 
\begin{equation}\label{eq:dp-avg}
\theta_{t+1} = \theta_t - \frac{1}{m}(\sum _{k=1}^K \Delta \theta_{t+1}^k + \mathcal{N}(0, \sigma^2))
\end{equation}

Our proposed attentive federated aggregation can also add this mechanism smoothly using Equation \ref{eq:dp}. The randomization is added in before the clients send the updated parameters to the server, but it is written in the form of server optimization for simplicity.
\begin{equation}\label{eq:dp}
\theta_{t+1} \gets \theta_t - \epsilon \sum_{k=1}^m\alpha_k(\theta_t-\theta_{t+1}^k + \beta \mathcal{N}(0, \sigma^2))
\end{equation}
In practice, we add a magnitude coefficient $\beta \in (0, 1]$ on the randomization of normal noise to control the effect of the randomization mechanism on the performance of federated aggregation. 

\section{Experiments}
\label{sec:exp}
This section describes the experiments conducted to evaluate our proposed method. Two baseline methods are compared and additional exploratory experiments are conducted to further the exploration of the performance of our attentive method.
Our code is available online at \url{https://github.com/shaoxiongji/fed-att}. 

\subsection{Datasets}
\label{sec::datasets}
We conduct experiments of neural language modeling experiments on three English language datasets for evaluation to mimic the real-world scenario of mobile keyboards in the decentralized applications. They are Penn Treebank \cite{marcus1993building}, WikiText-2 \cite{merity2016pointer}, and the Reddit Comments from Kaggle.
Language modeling is one of the most suitable tasks for the validation of federated learning. It has a large number of datasets to test the performance and there is a real-world application, i.e., the input keyboard application in smart phones. 

Penn Treebank is an annotated English corpus. We use the data derived from Zaremba et al.\footnote{Penn Treebank is available at \url{https://github.com/wojzaremba/lstm/tree/master/data}} \cite{zaremba2014recurrent}.
The WikiText-2 is available online\footnote{WikiText-2 is available at \url{https://s3.amazonaws.com/research.metamind.io/wikitext/wikitext-2-v1.zip}}.
May 2015 Reddit Comments dataset is a portion of a large scale dataset of Reddit comments\footnote{Avaliable at \url{https://www.reddit.com/r/datasets/comments/3bxlg7/i_have_every_publicly_available_reddit_comment/}, retrieved in Dec, 2018} from the popular online community -- Reddit. It is available in the Kaggle Datasets\footnote{Reddit Comments dataset is available at \url{https://www.kaggle.com/reddit/reddit-comments-may-2015}}. We sampled 1\textperthousand~of the comments from this dataset to train our private language model as a representative of social networks data. The statistical information, i.e., the number of tokens in the training, validation, and testing set of these three datasets is shown in Table \ref{tab:datasets}. 

\begin{table}[htbp]
\caption{Number of tokens in training, validation and testing sets of three datasets}
\small
\begin{center}
\label{tab:datasets}
\begin{tabular}{|c|c|c|c| } 
\toprule
Dataset & \# Train &  \# Valid. & \# Test\\ 
\midrule
Penn Treebank & 887,521 & 70,390  & 78,669 \\ 
WikiText-2 & 2,088,628 & 217,646  & 245,569 \\ 
Reddit Comments & 1,784,023 & 102,862   & 97,940 \\ 
\bottomrule
\end{tabular}
\end{center}
\end{table}

\noindent \textbf{Data~Partitioning.~} To mimic the scenario of real-world private keyboard applications, we perform data partitioning on these three popular language modeling datasets. At first, we shuffle the whole dataset. Then, we perform random sampling without replacement under the independently identical distribution. The whole dataset is partitioned into a certain number of shards denoted as the number of users or clients. We split these three datasets into 100 subsets as the user groups of 100 clients to participate in the federated aggregation after local training.

\subsection{Baselines~and~Settings}
\label{exp::baselines}
We conducted to several groups experiments for comparison, for example, performance with different model aggregation methods, the scale of client models, communication cost, and so forth. There are two baselines totally in these comparisons, i.e., \texttt{FedSGD} and \texttt{FedAvg}. The basic definitions and settings of baselines and our proposed method are as follows.

\begin{enumerate}
\item \texttt{FedSGD}: Federated stochastic gradient descent takes all the clients for federated aggregation and each client performs one epoch of gradient descent.
\item \texttt{FedAvg}: Federated averaging samples a fraction of users for each iteration and each client can take several steps of gradient descent.
\item \texttt{FedAtt}: Our proposed FedAtt takes a similar setting as FedAvg, but uses an improved attentive aggregation algorithm.
\end{enumerate}

We conduct experiments under the setting of federated learning using the GRU-based private neural language modeling with Nvidia GTX 1080 Ti GPU acceleration. The GRU-based client model firstly takes texts as input, then embeds them into word vectors and feeds them to the GRU network. The last fully connected layer takes the output of GRU as input to predict the next word. The small model uses 300 dimensional word embedding and hidden state of RNN unit. We deploy models of three scales: small, medium and large with word embedding dimensions of 300, 650 and 1500, respectively. Tied embedding is applied to reduce the size of the model and decrease the communication cost. Tied embedding shares the weights and biases in the embedding layer and output layer and greatly reduces the number of trainable parameters greatly.

\subsection{Results}
\label{exp:lang}
We conduct experiments on these three datasets and three federated learning methods. Testing perplexity is taken as the evaluation metric. Perplexity is a standard measurement for probability distribution. It is one of the most commonly used metrics for word-level language modeling. The perplexity of a distribution is defined as 
\begin{equation*}
PPL(x)=2^{H(p)}=2^{-\sum_xp(x)\log \frac{1}{p(x)}}
\end{equation*}
where $H(p)$ is the entropy of the distribution $p(x)$. A lower perplexity stands for a better prediction performance of the language model.

We take 50 rounds of communication between server and clients and compare the performance on the validation set to select the best model, then test the performance on the testing set to get the testing perplexity. The results of testing perplexity of all three datasets are shown in Table \ref{exp:ppl}. For FedAvg and FedAtt, we set the client fraction $C$ to be 0.1 and 0.5 within these results. According to the definition of FedSGD, the client fraction is always 1. 
As shown in this table, our proposed FedAtt outperforms FedSGD and FedAvg in terms of testing perplexity in all the three datasets. When the client fraction $C$ is 0.1 and 0.5 in the Penn Treebank and WikiText-2 respectively, our proposed method gains a significant improvement over its counterparts. We also conduct experiments on the fine-grained setting of the client fraction $C$ (from 0.1 to 0.9). When the client fraction is 0.7, our proposed method obtains the best testing perplexity of 67.59 in the WikiText-2 dataset.

\begin{table}[htp]
\caption{Testing perplexity of 50 communication rounds for federated training using small-scaled GRU network as the client model}
\small
  \begin{center}
	\begin{tabular}{|c|c|c|c|c|}
	\toprule
	Frac. & Methods & WikiText-2 & PTB & Reddit \\
	\midrule
	1 & FedSGD &112.45 &155.27 &128.61 \\
	\hline
	\multirow{2}{1em}{0.1} & FedAvg & 95.27 &138.13 & 126.49\\
	 & FedAtt & 91.82 &115.43 & 120.25\\
	\hline
	\multirow{2}{1em}{0.5} & FedAvg & 79.75  & 128.24 & 101.64 \\
	 & FedAtt & 69.38  & 123.00 & 99.04\\
	\midrule
	\end{tabular}
  \end{center}
  \label{exp:ppl}
\end{table}

We then further our exploration of the four factors in the WikiText-2 dataset to evaluate the performance of our proposed method with a comparison of its counterpart FedAvg. In additional exploratory experiments in the following subsections, we explored the client fraction, the communication costs, the effect of different randomizations, and the scale of the models.

\subsection{Client~Fraction}
\begin{figure}[htbp]
\begin{center}
\includegraphics[width=0.49\textwidth]{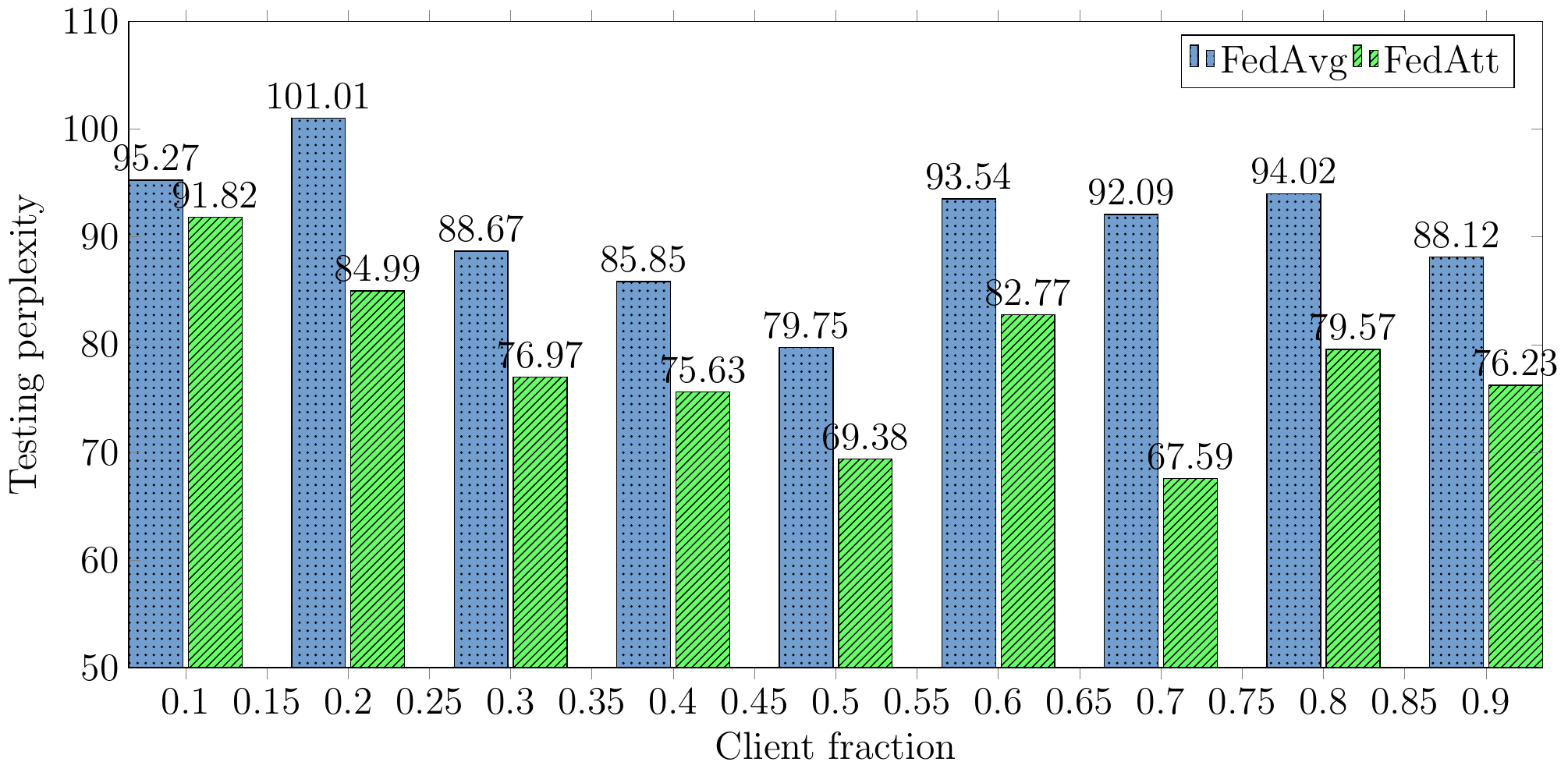}
\caption{Testing perplexity of 50 communication rounds when a different number of clients is selected for federated aggregation}
\label{exp:ppl-frac}
\end{center}
\end{figure}

In real-world applications of federated learning, some clients may be offline due to a change in user behavior or network issues. Thus, it is necessary to choose only a small number of clients for federated optimization. To evaluate the effect of the client fraction $C$ on the performance of our proposed attentive federated optimization, we explore the testing perplexity with various number of clients. The result is illustrated in Figure \ref{exp:ppl-frac} where the client fraction varies from 0.1 to 0.9. The small-scaled neural language model is used in this evaluation. The testing perplexity fluctuates when the client fraction increases. There is no guarantee that more clients results in a better score. Actually, 70\% of clients for model aggregation achieved the lowest perplexity in this experiment. This result indicates that the number of clients participating in model aggregation has an impact on the performance. But our proposed FedAtt can achieve much quite lower perplexity than FedAvg for all the settings of the client fraction.

\subsection{Communication~Cost}
Communication cost for parameter uploading and downloading between the clients and server is another important issue for decentralized learning. Communication, both wired and wireless, depends on Internet bandwidth highly and has an impact on the performance of federated optimization. To save the capacity of network communication, decentralized training should be more communication-efficient. 
Several approaches apply compression methods to achieve efficient communication. Our method accelerates the training through the optimization of the global server as it can converge more quickly than its counterparts. 

\begin{figure}[htbp]
\begin{center}
  \begin{subfigure}[]{0.47\textwidth}
  	\includegraphics[width=\textwidth]{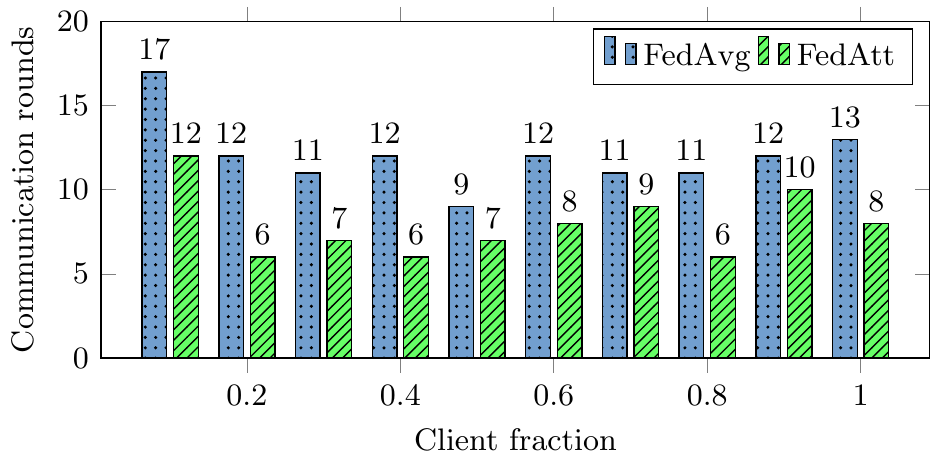} 
	\label{fig:rounds-frac}
	\subcaption{Rounds vs. client fraction}
  \end{subfigure}
  \quad
  \begin{subfigure}[]{0.23\textwidth}
  	\includegraphics[width=\textwidth]{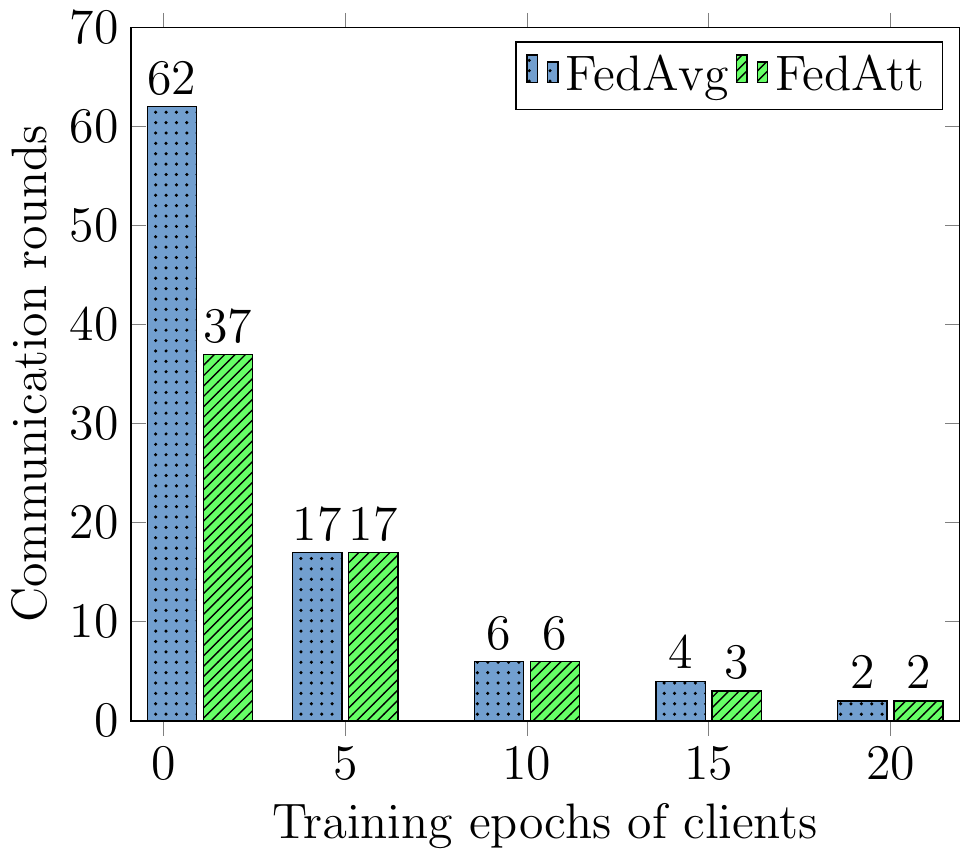} 
	\label{fig:rounds-ep}
	\subcaption{Rounds vs. epochs}
  \end{subfigure}
  \quad
  \begin{subfigure}[]{0.23\textwidth}
  	\includegraphics[width=\textwidth]{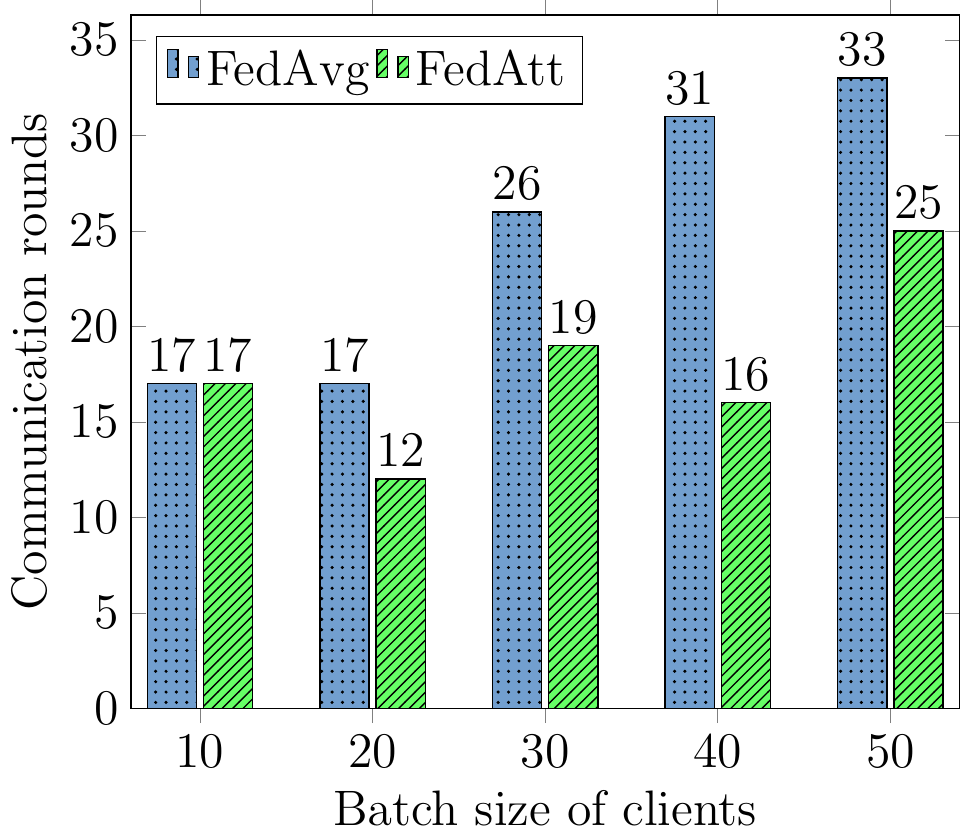}
	\label{fig:rounds-bs}
	\subcaption{Rounds vs. batch size}
  \end{subfigure}
\caption{Effect of the client fraction, epochs, and batch size of clients on communication rounds when the threshold of testing perplexity is set to be 90 and small-scaled GRU-based language model is used}
\label{exp:rounds}
\end{center}
\end{figure}

To compare the efficiency of communication, we take the communication rounds during training as the evaluation metric in this subsection. 
Three factors are considered, i.e., the client fraction, epochs and batch size of client training. The results are shown in Figure \ref{exp:rounds} where the small-scaled language model is used as the client model and 10\% of clients are selected for model aggregation. We set the testing perplexity for the termination of federated training to be 90. When the testing perplexity is lower than that threshold, federated training comes to an end and we take the rounds of training as the communication rounds. As shown in Figure \ref{exp:rounds}(a), the communication round during training fluctuates when the number of client increases. Furthermore, our proposed method is always better than FedAvg with less communication cost. When the client fraction $C$ chosen is 0.2 and 0.4, our proposed method saves a half of communication rounds.
Then, we evaluate the effect of the local computation of clients on the communication rounds. We take the local training epochs to be 1, 5, 10, 15, and 20 and the local batch size to be from 10 to 50. We proposed FedAtt to achieve a comparable communication cost in the comparison of different values of epoch and the batch size of local training, as shown in Figure \ref{exp:rounds}(b) and Figure \ref{exp:rounds}(c) respectively. 

\subsection{Magnitude~of~Randomization}
The federated learning framework focuses on the privacy of the input data using distributed training on each client side to protect the user's privacy. To further the privacy preservation of the decentralized training, we evaluate the magnitude of normal noise in the randomization mechanism on model parameters. Comparative experiments are conducted to analyze the effect of the magnitude on the testing perplexity.  The results are shown in Table \ref{exp:dp} with both randomized and nonrandomized settings. For the randomized version, four values of magnitude are chosen, i.e., 0.001, 0.005, 0.01, and 0.05. 

As shown in the table, a very small noise on both the methods does not affect the performance. Actually, the testing perplexity for the randomized setting is slightly better than the result of nonrandomized setting.  With a larger noise, the performance becomes worse. For our proposed method, the testing perplexity is always lower than its counterpart FedAvg, showing that our method can resist a larger noise and can better preserve privacy to ensure the perplexity of next-word prediction.

\begin{table}[htp]
\caption{Magnitude of randomization vs. testing perplexity using a small-scaled model with tied embedding}
\small
\begin{center}
\begin{tabular}{|c|c|c|c|}
\toprule 
\multicolumn{2}{|c|}{Randomization} & FedAvg & FedAtt\\
\midrule
Nonrandomized & $\beta=0$ & 88.21 &77.66 \\
\hline
\multirow{4}{4em}{Randomized}& $\beta=0.001$& 88.17 & 77.76\\
&$\beta=0.005$ & 88.36 &78.59\\
&$\beta=0.01$ & 89.74& 79.51\\
&$\beta=0.05$ & 103.17 &101.82 \\
\bottomrule 
\end{tabular}
\end{center}
\label{exp:dp}
\end{table}%

\subsection{Scale~of~Model}
Distributed training depends on communication between the server and clients, and the central server needs to optimize on the model parameters for the aggregation of the clients models. Thus, the central server will have a higher communication cost and computational cost when there are a larger number of clients and the local models have millions of parameters.

The size of the vocabulary in most language modeling corpus is very large. To save training costs, the embedding weights and output weights are tied which can reduce the number of trainable parameters \cite{inan2016tying, press2017using}. We compared three scales of client models with the word embedding dimensions of 300, 650 and 1500. Two versions of the tied and untied models are used. In the tied setting, the dimension of the RNN hidden state must be the same as the embedding dimension. 

The results of the model's scales on the testing perplexity are shown in Table \ref{exp:scale}. The tied large-scale model achieves the best results for both FedAvg and FedAtt and the tied model is better than the untied model of the same scale. Our proposed method achieves lower testing perplexity in four out of the six settings, i.e., tied and untied small model, tied medium model, and tied large model. For the other two settings, the testing perplexity of our method is slightly higher than FedAvg. Overall, for real-world keyboard applications in practice, the tied embedding can be used to save the number of trainable parameters and the communication cost while achieving a better performance.

\begin{table}[htp]
\caption{Testing perplexity of 50 communication rounds vs. the scale of the model using a tied embedding or untied embedding model}
\small
\begin{center}
\begin{tabular}{|c|c|c|c|}
\toprule
\multicolumn{2}{|c|}{Model} & FedAvg & FedAtt \\
\toprule
\hline
\multirow{2}{3em}{Small} & tied & 88.21 &77.66 \\
\cline{2-4}
 & untied &91.25 &81.31\\
 \hline
 \hline
\multirow{2}{3em}{Medium}  &tied& 103.07 &77.41\\
\cline{2-4}
 & untied & 96.67 & 96.71 \\
 \hline
 \hline
\multirow{2}{3em}{Large}  &tied&77.51& 76.37 \\
\cline{2-4}
 & untied &82.97&83.40\\
\bottomrule
\end{tabular}
\end{center}
\label{exp:scale}
\end{table}%

\section{Conclusion}\label{sec:conclusion}
Federated learning provides a promising and practical approach to learning from decentralized data while protecting the private data with differential privacy. Efficient decentralized learning is significant for distributed real-world applications such as personalized keyboard word suggestion on mobile phones, providing a better service and protect user's private personal data.

To optimize the server aggregation by federated averaging, we investigated the model aggregation and optimization on the central server in this paper. We proposed a novel layer-wise attentive federated optimization for private neural language modeling which can measure the importance of selected clients and accelerate the learning process. We partitioned three popular datasets, i.e., Penn Treebank and WikiText-2 for the prototypical language modeling task, and Reddit comments from a real-world social networking website, to mimic the scenario of word-level keyboard suggestions and performed a series of exploratory experiments. Experiments on these datasets show our proposed method outperforms its counterparts in most settings. 

\section*{Acknowledgement}
This research is funded by the Australian Government through the Australian Research Council (ARC) under grants LP150100671 partnership with Australia Research Alliance for Children and Youth (ARACY) and Global Business College Australia (GBCA).

\bibliographystyle{unsrt}
\bibliography{cite_fedatt}

\begin{thebibliography}{10}

\bibitem{mcmahan2017communication}
Brendan McMahan, Eider Moore, Daniel Ramage, Seth Hampson, and Blaise~Aguera
  y~Arcas.
\newblock Communication-efficient learning of deep networks from decentralized
  data.
\newblock In {\em Artificial Intelligence and Statistics}, pages 1273--1282,
  2017.

\bibitem{geyer2017differentially}
Robin~C Geyer, Tassilo Klein, and Moin Nabi.
\newblock Differentially private federated learning: A client level
  perspective.
\newblock {\em arXiv preprint arXiv:1712.07557}, 2017.

\bibitem{chen2018federated}
Fei Chen, Zhenhua Dong, Zhenguo Li, and Xiuqiang He.
\newblock Federated meta-learning for recommendation.
\newblock {\em arXiv preprint arXiv:1802.07876}, 2018.

\bibitem{kim2016predicting}
Eunice Kim, Jung-Ah Lee, Yongjun Sung, and Sejung~Marina Choi.
\newblock Predicting selfie-posting behavior on social networking sites: An
  extension of theory of planned behavior.
\newblock {\em Computers in Human Behavior}, 62:116--123, 2016.

\bibitem{arnold2016suggesting}
Kenneth~C Arnold, Krzysztof~Z Gajos, and Adam~T Kalai.
\newblock On suggesting phrases vs. predicting words for mobile text
  composition.
\newblock In {\em Proceedings of the 29th Annual Symposium on User Interface
  Software and Technology}, pages 603--608. ACM, 2016.

\bibitem{he2017new}
Hongmei He, Tim Watson, Carsten Maple, J{\"o}rn Mehnen, and Ashutosh Tiwari.
\newblock A new semantic attribute deep learning with a linguistic attribute
  hierarchy for spam detection.
\newblock In {\em 2017 International Joint Conference on Neural Networks
  (IJCNN)}, pages 3862--3869. IEEE, 2017.

\bibitem{hochreiter1997long}
Sepp Hochreiter and J{\"u}rgen Schmidhuber.
\newblock Long short-term memory.
\newblock {\em Neural computation}, 9(8):1735--1780, 1997.

\bibitem{cho2014learning}
Kyunghyun Cho, Bart van Merrienboer, Caglar Gulcehre, Dzmitry Bahdanau, Fethi
  Bougares, Holger Schwenk, and Yoshua Bengio.
\newblock Learning phrase representations using rnn encoder--decoder for
  statistical machine translation.
\newblock In {\em Proceedings of the 2014 Conference on Empirical Methods in
  Natural Language Processing (EMNLP)}, pages 1724--1734, 2014.

\bibitem{popov2018distributed}
Vadim Popov, Mikhail Kudinov, Irina Piontkovskaya, Petr Vytovtov, and Alex
  Nevidomsky.
\newblock Distributed fine-tuning of language models on private data.
\newblock In {\em International Conference on Learning Representation (ICLR)},
  2018.

\bibitem{kim2017federated}
Yejin Kim, Jimeng Sun, Hwanjo Yu, and Xiaoqian Jiang.
\newblock Federated tensor factorization for computational phenotyping.
\newblock In {\em Proceedings of the 23rd ACM SIGKDD International Conference
  on Knowledge Discovery and Data Mining}, pages 887--895. ACM, 2017.

\bibitem{smith2017federated}
Virginia Smith, Chao-Kai Chiang, Maziar Sanjabi, and Ameet~S Talwalkar.
\newblock Federated multi-task learning.
\newblock In {\em Advances in Neural Information Processing Systems}, pages
  4427--4437, 2017.

\bibitem{konevcny2016federated}
Jakub Kone{\v{c}}n{\`y}, H~Brendan McMahan, Felix~X Yu, Peter Richt{\'a}rik,
  Ananda~Theertha Suresh, and Dave Bacon.
\newblock Federated learning: Strategies for improving communication
  efficiency.
\newblock {\em arXiv preprint arXiv:1610.05492}, 2016.

\bibitem{alistarh2017qsgd}
Dan Alistarh, Demjan Grubic, Jerry Li, Ryota Tomioka, and Milan Vojnovic.
\newblock {QSGD}: Communication-efficient {SGD} via gradient quantization and
  encoding.
\newblock In {\em Advances in Neural Information Processing Systems}, pages
  1709--1720, 2017.

\bibitem{wen2017terngrad}
Wei Wen, Cong Xu, Feng Yan, Chunpeng Wu, Yandan Wang, Yiran Chen, and Hai Li.
\newblock Terngrad: Ternary gradients to reduce communication in distributed
  deep learning.
\newblock In {\em Advances in Neural Information Processing Systems}, pages
  1509--1519, 2017.

\bibitem{mikolov2010recurrent}
Tom{\'a}{\v{s}} Mikolov, Martin Karafi{\'a}t, Luk{\'a}{\v{s}} Burget, Jan
  {\v{C}}ernock{\`y}, and Sanjeev Khudanpur.
\newblock Recurrent neural network based language model.
\newblock In {\em Eleventh Annual Conference of the International Speech
  Communication Association}, 2010.

\bibitem{jozefowicz2016exploring}
Rafal Jozefowicz, Oriol Vinyals, Mike Schuster, Noam Shazeer, and Yonghui Wu.
\newblock Exploring the limits of language modeling.
\newblock {\em arXiv preprint arXiv:1602.02410}, 2016.

\bibitem{inan2016tying}
Hakan Inan, Khashayar Khosravi, and Richard Socher.
\newblock Tying word vectors and word classifiers: A loss framework for
  language modeling.
\newblock {\em arXiv preprint arXiv:1611.01462}, 2016.

\bibitem{press2017using}
Ofir Press and Lior Wolf.
\newblock Using the output embedding to improve language models.
\newblock In {\em Proceedings of the 15th Conference of the European Chapter of
  the Association for Computational Linguistics: Volume 2, Short Papers},
  volume~2, pages 157--163, 2017.

\bibitem{mnih2014recurrent}
Volodymyr Mnih, Nicolas Heess, Alex Graves, et~al.
\newblock Recurrent models of visual attention.
\newblock In {\em Advances in neural information processing systems}, pages
  2204--2212, 2014.

\bibitem{bahdanau2014neural}
Dzmitry Bahdanau, Kyunghyun Cho, and Yoshua Bengio.
\newblock Neural machine translation by jointly learning to align and
  translate.
\newblock {\em arXiv preprint arXiv:1409.0473}, 2014.

\bibitem{luong2015effective}
Thang Luong, Hieu Pham, and Christopher~D Manning.
\newblock Effective approaches to attention-based neural machine translation.
\newblock In {\em Proceedings of the 2015 Conference on Empirical Methods in
  Natural Language Processing}, pages 1412--1421, 2015.

\bibitem{yin2016abcnn}
Wenpeng Yin, Hinrich Sch{\"u}tze, Bing Xiang, and Bowen Zhou.
\newblock Abcnn: Attention-based convolutional neural network for modeling
  sentence pairs.
\newblock {\em Transactions of the Association of Computational Linguistics},
  4(1):259--272, 2016.

\bibitem{yang2016hierarchical}
Zichao Yang, Diyi Yang, Chris Dyer, Xiaodong He, Alex Smola, and Eduard Hovy.
\newblock Hierarchical attention networks for document classification.
\newblock In {\em Proceedings of the 2016 Conference of the North American
  Chapter of the Association for Computational Linguistics: Human Language
  Technologies}, pages 1480--1489, 2016.

\bibitem{shen2017disan}
Tao Shen, Tianyi Zhou, Guodong Long, Jing Jiang, Shirui Pan, and Chengqi Zhang.
\newblock Disan: Directional self-attention network for rnn/cnn-free language
  understanding.
\newblock {\em arXiv preprint arXiv:1709.04696}, 2017.

\bibitem{finn2017model}
Chelsea Finn, Pieter Abbeel, and Sergey Levine.
\newblock Model-agnostic meta-learning for fast adaptation of deep networks.
\newblock {\em arXiv preprint arXiv:1703.03400}, 2017.

\bibitem{nichol2018firstorder}
Alex Nichol, Joshua Achiam, and John Schulman.
\newblock On first-order meta-learning algorithms.
\newblock {\em arXiv preprint arXiv:1803.02999}, 2018.

\bibitem{pan2010survey}
Sinno~Jialin Pan and Qiang Yang.
\newblock A survey on transfer learning.
\newblock {\em IEEE Transactions on Knowledge and Data Engineering},
  22(10):1345--1359, 2010.

\bibitem{marcus1993building}
Mitchell~P Marcus, Mary~Ann Marcinkiewicz, and Beatrice Santorini.
\newblock Building a large annotated corpus of english: The {P}enn {T}reebank.
\newblock {\em Computational linguistics}, 19(2):313--330, 1993.

\bibitem{merity2016pointer}
Stephen Merity, Caiming Xiong, James Bradbury, and Richard Socher.
\newblock Pointer sentinel mixture models.
\newblock {\em arXiv preprint arXiv:1609.07843}, 2016.

\bibitem{zaremba2014recurrent}
Wojciech Zaremba, Ilya Sutskever, and Oriol Vinyals.
\newblock Recurrent neural network regularization.
\newblock {\em arXiv preprint arXiv:1409.2329}, 2014.

\end{thebibliography}
\end{document}